\RequirePackage{fix-cm}
\documentclass[twocolumn]{svjour3}          
\smartqed  
\usepackage{graphicx}
\usepackage{times}
\usepackage{epsfig}
\usepackage{graphicx}
\usepackage{amsmath}
\usepackage{amssymb}
\usepackage{dsfont}
\usepackage{tabularx}
\usepackage{subcaption}
\usepackage{pifont} 
\usepackage{soul}
\newcommand{\cmark}{\ding{51}}%
\newcommand{\xmark}{\ding{55}}%
\usepackage[pagebackref=true,breaklinks=true,colorlinks,bookmarks=false]{hyperref}

\newcommand{\figref}[1]{Fig.~\ref{#1}}
\newcommand{\tabref}[1]{Tab.~\ref{#1}}
\newcommand{\secref}[1]{Sec.~\ref{#1}}

\usepackage{color}
\usepackage{hyperref}
\newcommand{\figAHuP}{
\begin{figure*}[t]
    \centering
    \includegraphics[width=.9\linewidth]{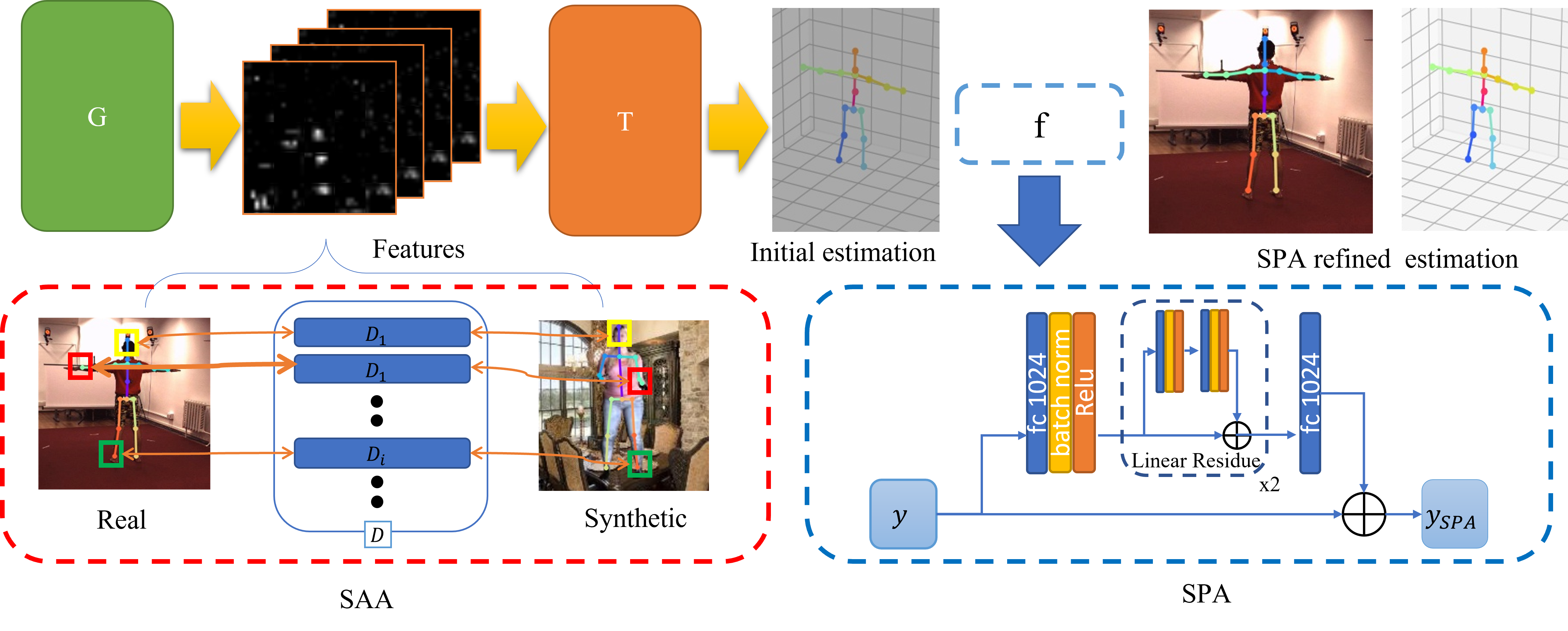}
    \caption{The proposed adapted human pose (AHuP) framework,
    where SAA stands for the semantically aware adaptation approach, SPA stands for the  skeletal  pose  adaptation. $G$ stands for the feature extractor, $T$ for task head, $D$ for discriminator and $D_i$ stands for the \textit{i}-th channel of $D$.}
    \label{fig:AHuP}
    \vspace{-.2in}
\end{figure*}
}

\newcommand{\figSAAandSPA}{
\begin{figure}
    \centering
    \includegraphics[width=0.9\linewidth]{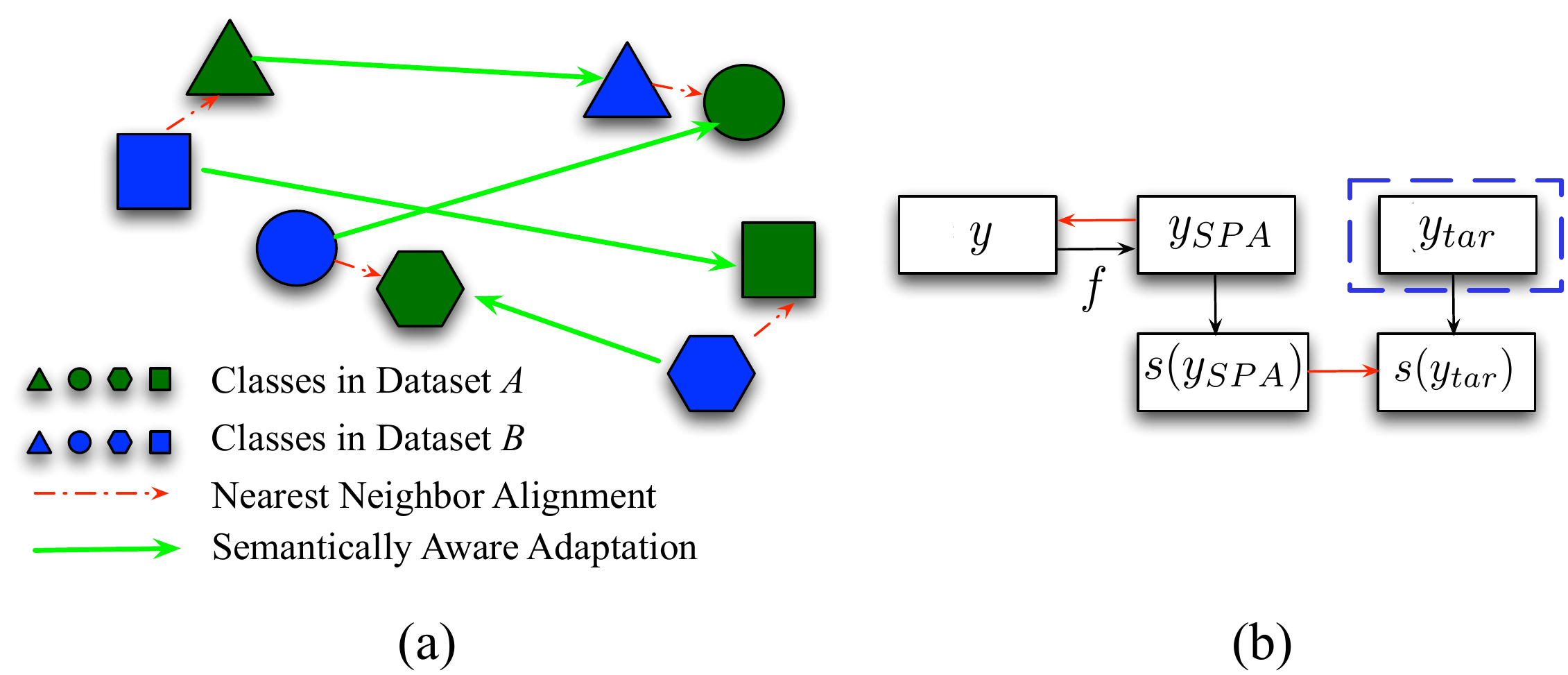}
    \caption{(a) Semantically aware adaptation (SAA) vs. nearest neighbor alignment between datasets $A$ and $B$, assumed to be from two different domains. (b) Skeletal pose adaptation (SPA) based on dual direction pivoting to both source and target in different representation. Black arrow indicates the forward mapping. Red arrow indicates the pivoting direction.}
    \label{fig:SAAandSPA}
    \vspace{-.2in}
\end{figure}
}

\newcommand{\figTSNEpose}{
\begin{figure}
    \centering
    \includegraphics[width=0.7\linewidth]{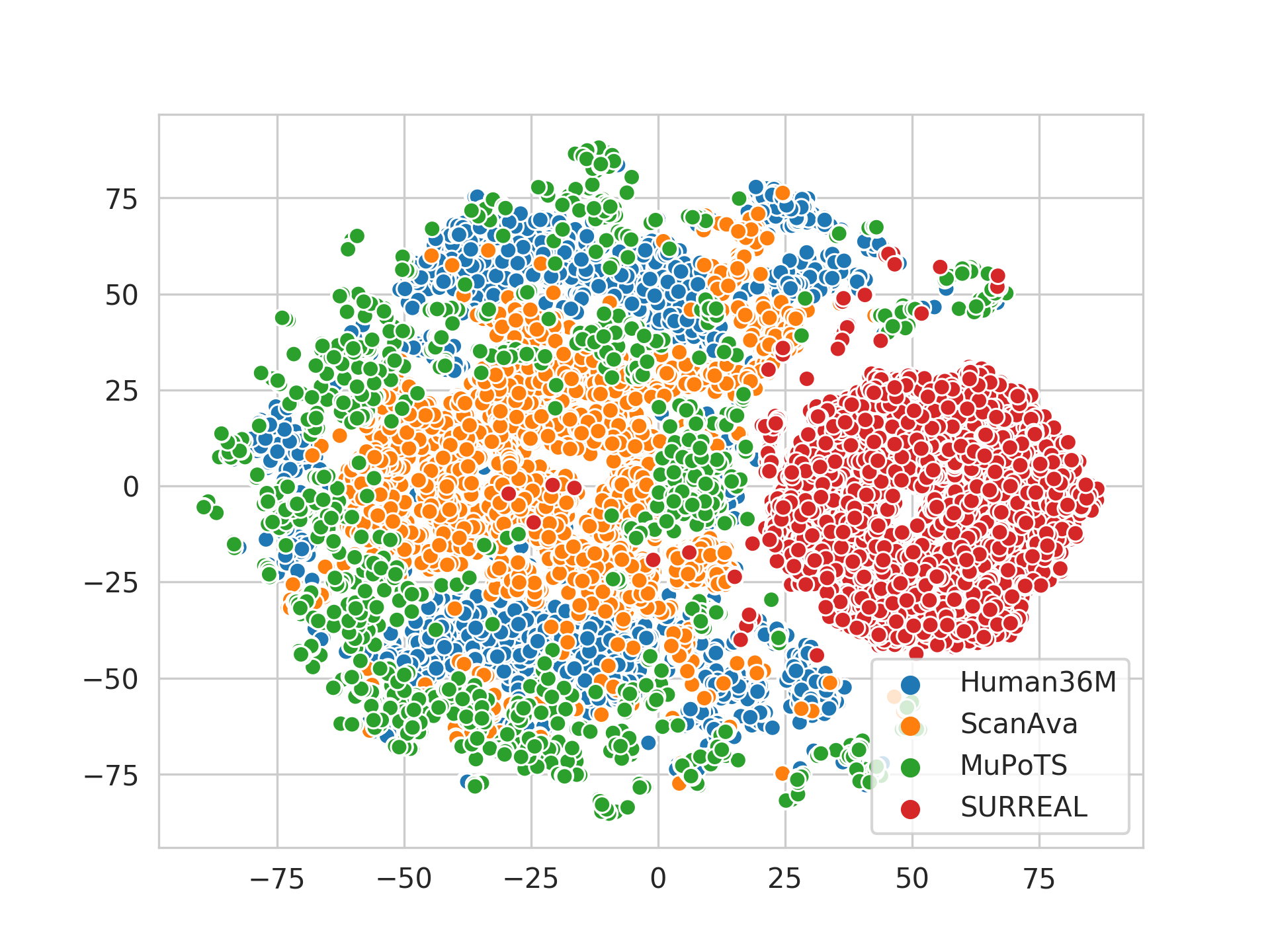}
    \caption{t-SNE plot of pelvis rooted 3D pose data across  5000 data samples randomly extracted from Human3.6M, MuPoTS, ScanAva+ and SURREAL. }
    \label{fig:tSNEpose}
     \vspace{-.25in}
\end{figure}
}

\newcommand{\figTSNEapr}{
\begin{figure*}[h]
    \centering
    \subfloat[]{\label{fig:D0}\includegraphics[width=0.25\linewidth]{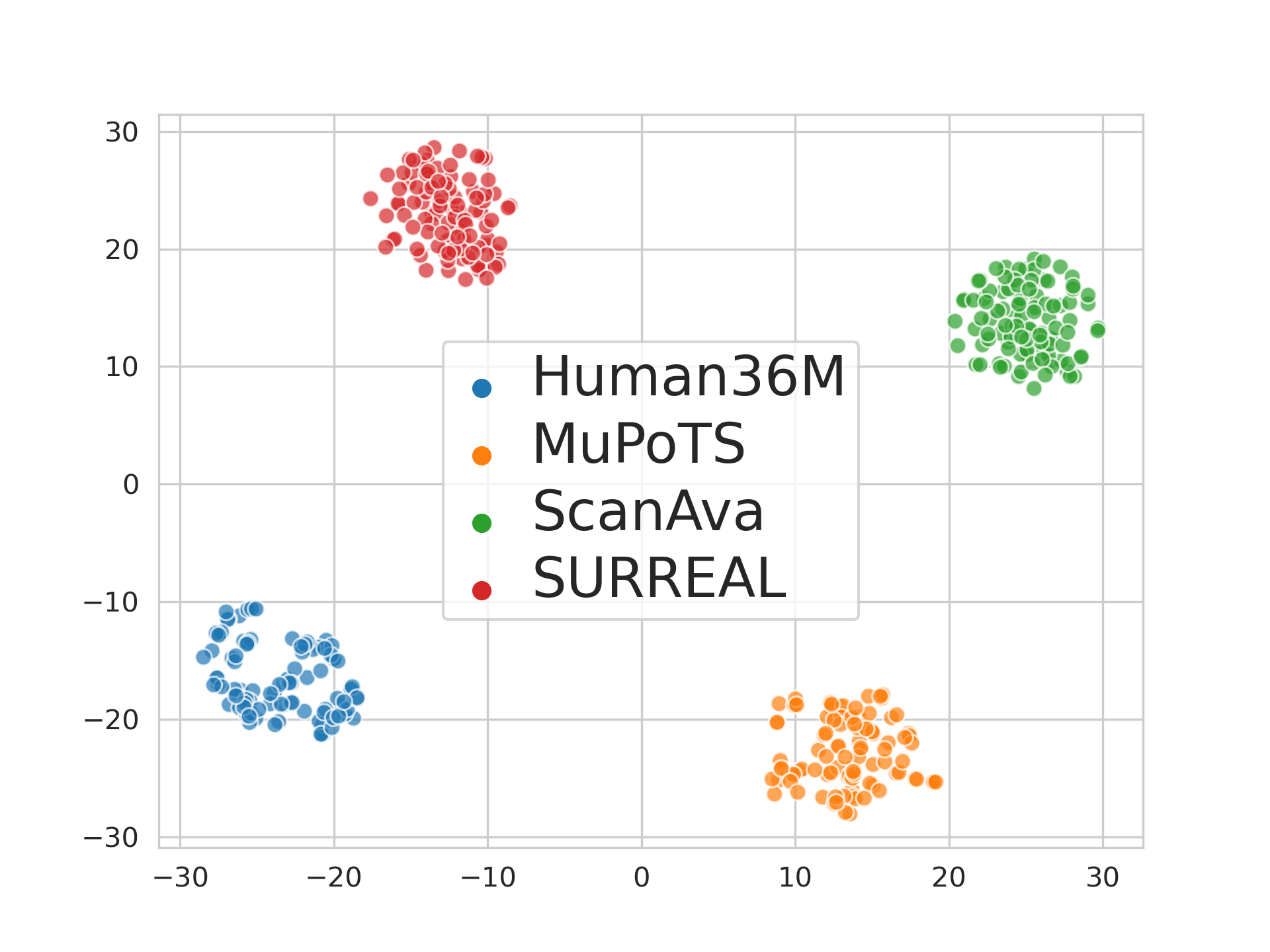}}
    \subfloat[]{\label{fig:C}\includegraphics[width=0.25\linewidth]{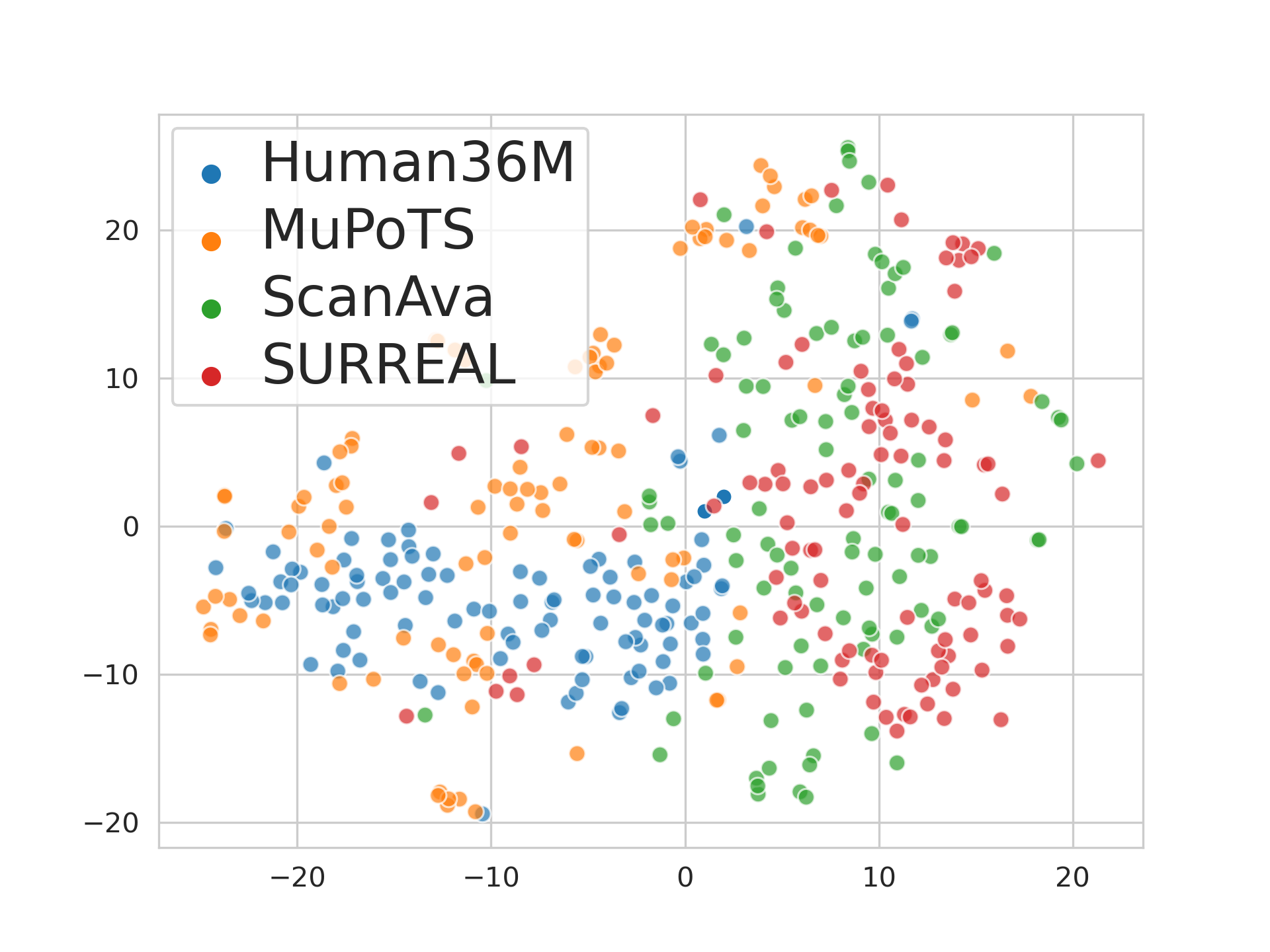}}
    \subfloat[]{\label{fig:SAA}\includegraphics[width=0.25\linewidth]{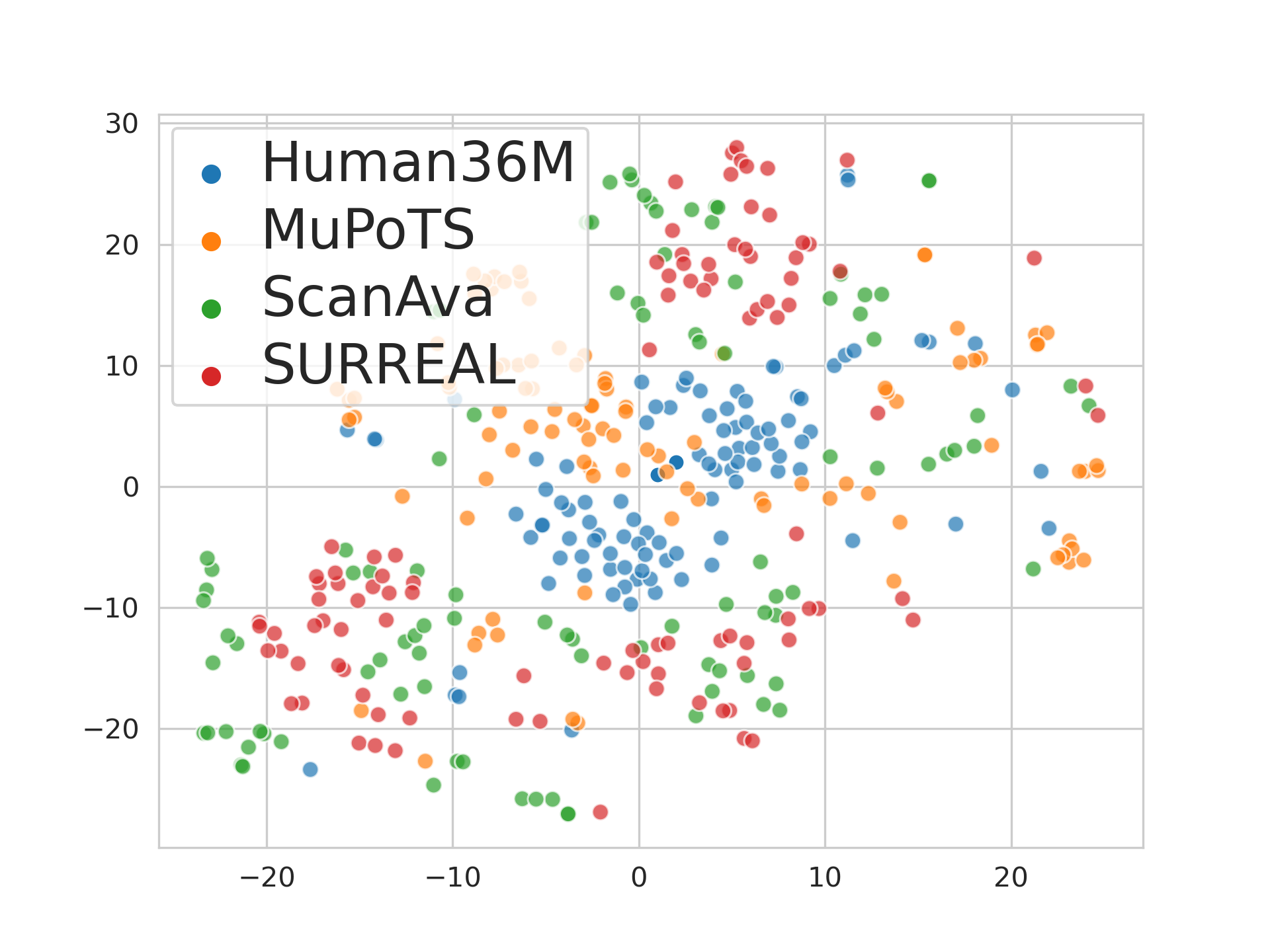}}
    \subfloat[]{\label{fig:SAA-y}\includegraphics[width=0.25\linewidth]{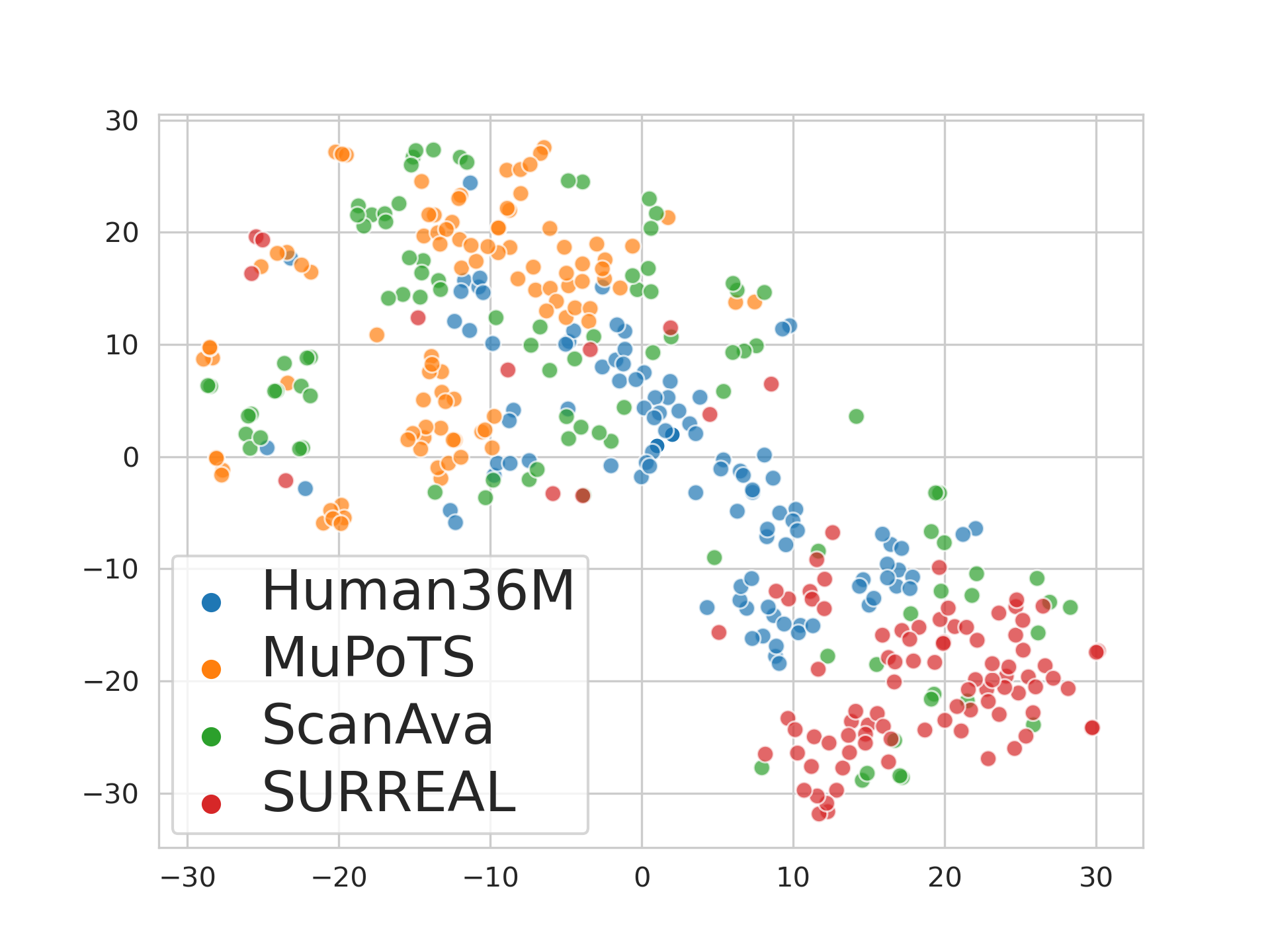}}

    \caption{t-SNE plot of network $G$'s output features applied on Human3.6M, MuPoTS, ScanAva+, and SURREAL datasets under model configurations of (a) no adaptation, (b) with a conventional adaptor C, (c) with SAA,  (d) with SAA + Jo2D.}
    \vspace{-.1in}
    \label{fig:tSNEapr}
\end{figure*}
}

\newcommand{\figSPAparams}{
\begin{figure}
    \centering
    \subfloat[]{\label{fig:SAPp1}\includegraphics[width=0.45\linewidth]{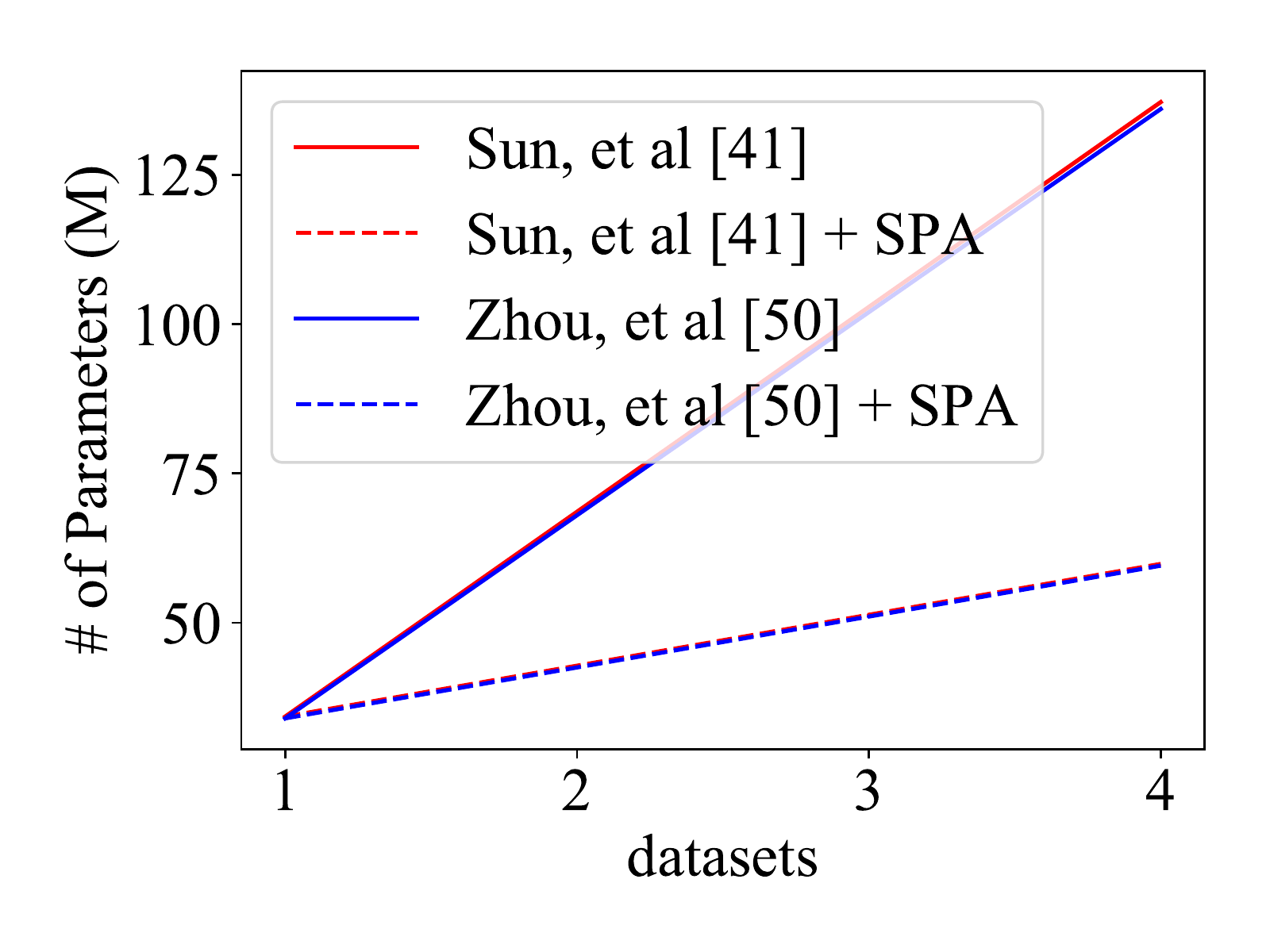}}
    \subfloat[]{\label{fig:SPAp2}\includegraphics[width=0.45\linewidth]{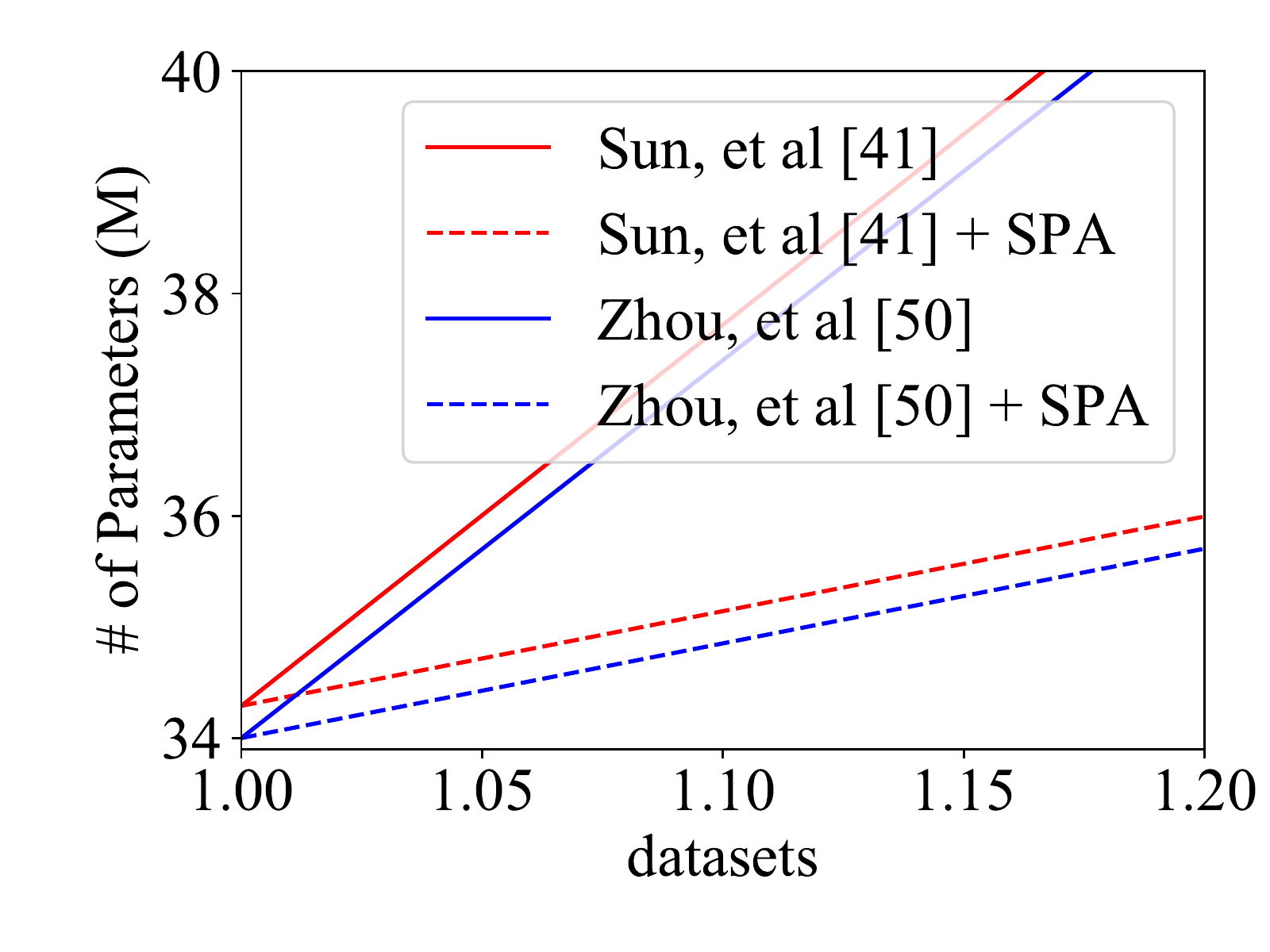}}
    \caption{Memory usage based on the  number of parameters in million (M) in the SOTA models from \cite{sun2018integral} and \cite{zhou2017towards}, when customized for varying numbers of datasets with and without SPA: (a) in full scale, (b) zoomed in to show the slope differences.}
    \label{fig:SPAparams}
\end{figure}
}

\newcommand{\figAAICvis}{
\begin{figure*}[h]
    \centering
    \includegraphics[width=0.99\linewidth]{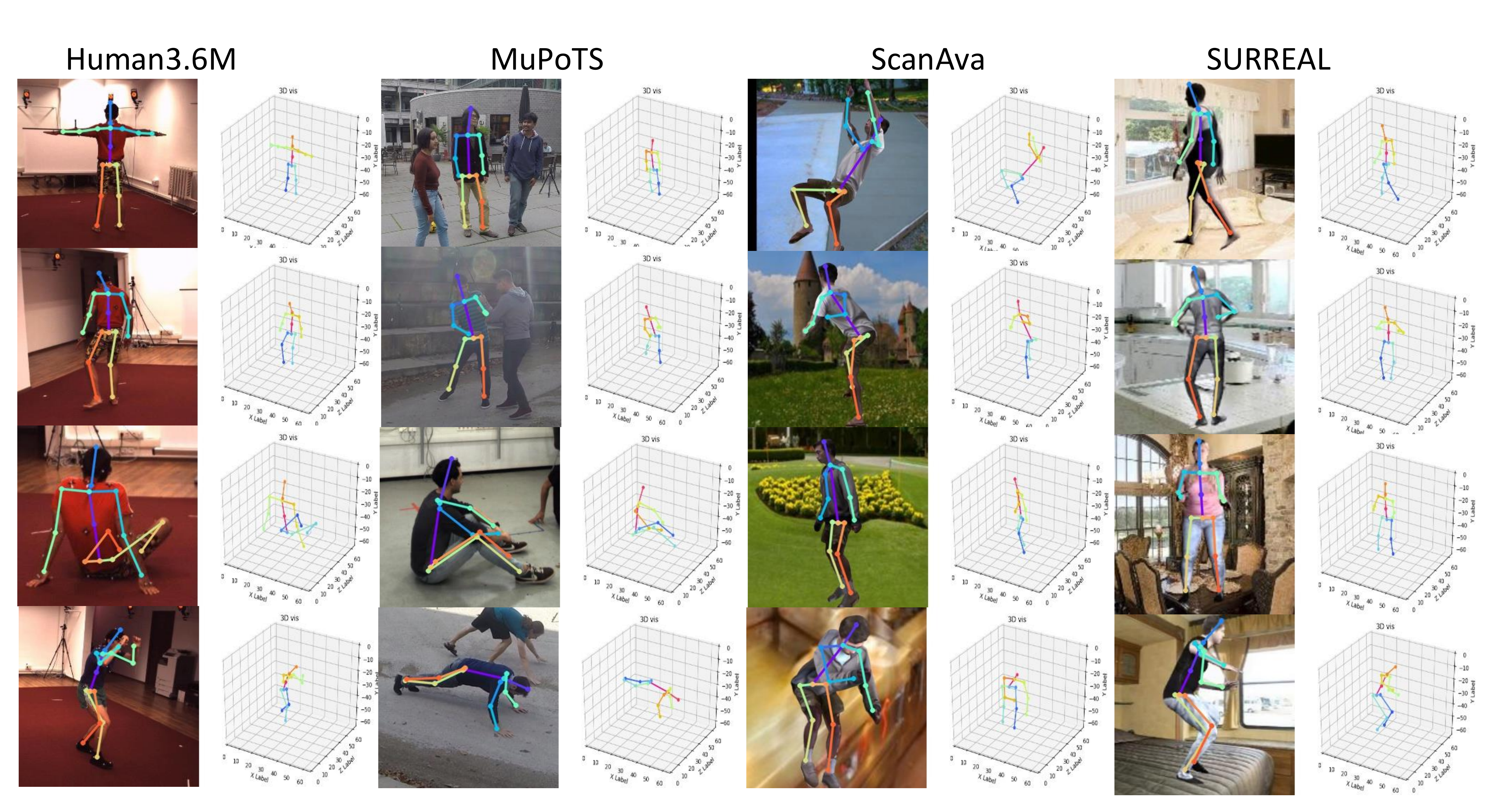}
    \caption{Qualitative recovery results of AHuP trained on ScanAva+ tested across domains and datasets.}
    \label{fig:AAICvis}
    \vspace{-.1in}
\end{figure*}
}

\newcommand{\figAAICmulti}{
\begin{figure}[h]
    \centering
    \includegraphics[width=0.8\linewidth]{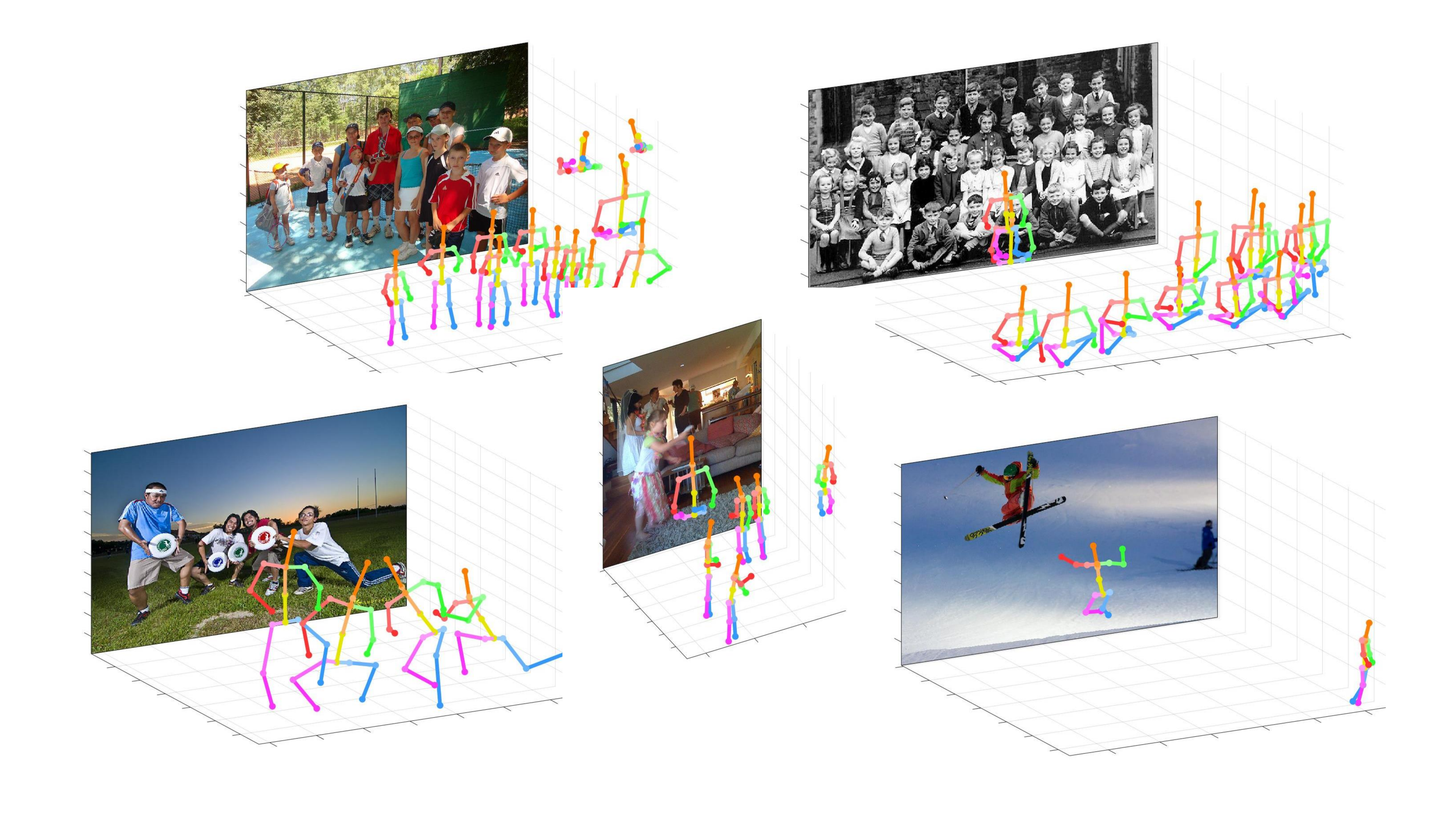}
    \caption{Qualitative recovery results of ScanAva-AHuP on MSCOCO dataset.}
    \label{fig:AAICmulti}
     \vspace{-.25in}
\end{figure}
}

\newcommand{\tblAbla}{
\begin{table*}[t]
\caption{Ablation study of AHuP when tested on real 3D human pose datasets as Human3.6M Protocol\#2 and MuPoTs. C stands for a conventional adaptation from \cite{chen2016synthesizing,long2015learning,sun2016deep}, SAA for semantic aware adaptation, Jo2D for  adding 2D pose estimation tasks from additional 2D human pose dataset (MSCOCO and MPII datasets), SPA for adding skeletal pose adaptation.}
 \vspace{-.2in}
\begin{center}
\footnotesize
 \begin{tabular}{c | *3c || *3c  }   
 \hline
  & \multicolumn{3}{c||}{Human3.6M Protocol\#2} &\multicolumn{3}{c}{MuPoTs}\\
  \hline
  Training Dataset + Adaptation Strategies  & PA MPJPE  & 3DPCK & AUC  & PA MPJPE  & 3DPCK & AUC  \\ 
    \hline 
ScanAva+ \cite{liu2018semi}   &  180.3 & 29.3 & 20.3  & 
            240.8 & 32.9 & 11.9 \\
 ScanAva+ + C \cite{chen2016synthesizing} &    156.5 & 58.8 & 24.4 &
           197.7 & 43.7  &  17.0 \\
 ScanAva+ + SAA &  99.9 & 84.0 & 41.7 &
            118.4 & 76.1 & 36.2 \\
 ScanAva+ + SAA + Jo2D &  88.8 & 89.0 & 45.9 & 
    111.8 & \textbf{81.8} & 38.7\\ 
 ScanAva + SAA + Jo2D + SPA & \textbf{85.1} & \textbf{90.2} & \textbf{47.9}  &
       \textbf{111.6} & 81.6 & \textbf{39.0}  \\ 
\hline
  \hline
  SURREAL \cite{varol17_surreal} & 181.4  &  44.9  & 16.4 &
                187.0 & 42.7 &  15.3 \\
 SURREAL + C \cite{chen2016synthesizing} &   145.5 & 39.7 & 27.8  & 
                165.1 & 55.5,& 21.7 \\
 SURREAL + SAA &  138.6 & 42.3 & 29.1 & 
        145.6  &  64.3  & 26.6 \\
  SURREAL + SAA +Jo2D &  135.5 &  46.2 & 30.3 & 
                134.6 & 71.3 & \textbf{30.0} \\
 SURREAL + SAA + Jo2D + SPA & \textbf{135.3} &  \textbf{68.8} & \textbf{30.3} & 
            \textbf{134.6}  &  \textbf{71.4} & 29.8\\
 \hline
\end{tabular}
\label{tbl:abla}
 \vspace{-.2in}
\end{center}
\end{table*}
}

\newcommand{\tblSPA}{
\begin{table}[b]

\caption{Cross benchmark evaluation of the 3D pose estimation models in \cite{sun2018integral} and \cite{zhou2017towards}, when trained and evaluated on different 3D pose datasets, with and without SPA. MuCo is the official training portion of the MuPoTS dataset \cite{singleshotmultiperson2018}.}
 \vspace{-.2in}
\begin{center}
\resizebox{.4\textwidth}{!}{
 \begin{tabular}{ *{5}c }   
 \hline
Benchmarks & \textbf{SPA} & MPJPE PA & 3DPCK & AUC \\
 \hline
 \multicolumn{5}{c}{Sun et al. \cite{sun2018integral} trained on MuCo + MSCOCO + MPII} \\
\hline 
Human36M & \xmark &  82.1 & 75.8 & 49.0 \\ 
    & \cmark &  \textbf{77.9} & \textbf{92.7} & \textbf{51.5} \\   
\hline  
ScanAva+ & \xmark &  92.6 & 87.7 & 43.7 \\ 
    & \cmark &  \textbf{91.7} & \textbf{87.7} & \textbf{44.2} \\   
\hline
SURREAL & \xmark & 154.8 & 63.0 & 26.6 \\ 
    & \cmark &  \textbf{154.2} & \textbf{63.2} & \textbf{27.0} \\   
\hline
\hline
 \multicolumn{5}{c}{Sun et al. \cite{sun2018integral} trained on Human3.6M + MSCOCO + MPII} \\ 
 \hline 
MuPoTS & \xmark & 111.1 & \textbf{84.1} & 41.0 \\ 
    & \cmark &  \textbf{105.9} & \textbf{84.1} & \textbf{41.1} \\   
\hline  
ScanAva+ & \xmark &  111.8 & 77.5 & 38.1 \\ 
    & \cmark &  \textbf{106.9} & \textbf{79.2} & \textbf{40.8} \\   
\hline
SURREAL & \xmark &  171.8 & 54.7 & 22.3 \\ 
    & \cmark &  \textbf{170.1} & \textbf{55.7} & \textbf{23.2} \\   
\hline
\hline
 \multicolumn{5}{c}{Zhou et al. \cite{zhou2017towards} trained on Human3.6M + MPII} \\ 
 \hline 
MuPoTS & \xmark & 111.8 & 77.5 & 38.1 \\ 
    & \cmark &  \textbf{106.9} & \textbf{79.2} & \textbf{40.8} \\   
\hline  
ScanAva+ & \xmark &  91.6 & 87.5 & 44.0 \\ 
    & \cmark &  \textbf{89.1} & \textbf{87.8} & \textbf{45.5} \\   
\hline
SURREAL & \xmark &  156.3 & 62.7 & 26.4 \\ 
    & \cmark &  \textbf{153.6} & \textbf{63.6} & \textbf{28.2} \\   
\hline

\end{tabular}
}
\label{tbl:SPA}
\end{center}
\end{table}
}

\newcommand{\tblSPAflops}{
\begin{table}[b]

\caption{Computational cost of the models in \cite{sun2018integral} and \cite{zhou2017towards} with and without SPA in Gigaflops.} 
 \vspace{-.2in}
\begin{center}
\resizebox{.4\textwidth}{!}{
 \begin{tabular}{ *{4}c }   
 \hline
Models & Original & +SPA & Computational cost increase (\%) \\
 \hline
 Sun et al. \cite{sun2018integral} &  13.097 & 13.101 & 0.031\%  \\
 Zhou et al. \cite{zhou2017towards} &  12.003 & 12.007 & 0.033\%  \\
 \hline
\end{tabular}
}
\label{tbl:SPAflops}
\end{center}
\end{table}
}

\newcommand{\tblHumanB}{
\begin{table*}[h]
\caption{Comparison with the state-of-the-art based on the MPJPE metric tested on Human3.6M dataset using Protocol\#2. Please note that unlike AHuP, all of these models have used some real 3D human pose data in their training process. For our method, PA stands for when the PA rigid alignment is applied. The last two rows stands for Martinez approach \cite{martinez2017simple} trained on ScanAva+ dataset.}
\vspace{-.1in}
\begin{center}
\small
\resizebox{\textwidth}{!}{
 \begin{tabular}{c  c   c    c     c  c   c  c   c    c     c   c   c  c   c    c     c  }   
 \hline
Methods  & Dir & Dis & Eat & Gre & Phon. & Pose & Pur. & Sit & SitD & Smo. & Phot. & Wait & Walk & WalkD. & WalkP. & Avg \\
\hline
Akhter \& Black \cite{akhter2015pose} & 
199.2 & 177.6 & 161.8 & 197.8 & 176.2 & 186.5&  195.4 &  167.3&  160.7&  173.7&  177.8&  181.9&  176.2&  198.6&  192.7&  181.1 \\
Ramakrishna \cite{ramakrishna2012reconstructing} & 
137.4 &  149.3 &  141.6 & 154.3 &  157.7 & 158.9 & 141.8 & 158.1 & 168.6 & 175.6 & 160.4 & 161.7 & 150.0 & 174.8 & 150.2 & 157.3 \\
Zhou \cite{zhou2016sparse} &
99.7 & 95.8 & 87.9 & 116.8 & 108.3 & 107.3 & 93.5 & 95.3 & 109.1 & 137.5 & 106.0 & 102.2 & 106.5 & 110.4 & 115.2 & 106.7 \\
SMPLify \cite{bogo2016keep} & 
62.0 & 60.2 & 67.8 & 76.5 & 92.1 & 77.0 & 73.0 & 75.3 & 100.3 & 137.3 & 83.4 & 77.3 & 79.7 & 86.8 & 81.7 & 82.3  \\ 
Chen\cite{chen20173d} & 89.9 & 97.6 & 90.0 & 107.9 & 107.3 & 93.6 & 136.1 & 133.1 & 240.1 & 106.7 & 139.2 & 106.2 & 87.0 & 114.1 & 90.6 & 114.2 \\
Tome \cite{tome2017lifting} & 65.0 & 73.5 & 76.8 & 86.4 & 86.3 & 68.9 & 74.8 & 110.2 & 173.9 & 85.0 & 110.7 & 85.8 & 71.4 & 86.3 & 73.1 & 88.4 \\
Moreno \cite{moreno20173d} & 69.5 & 80.2 & 78.2 & 87.0 & 100.8 & 76.0 & 69.7 & 104.7 & 113.9 & 89.7 & 102.7 & 98.5 & 79.2 & 82.4 & 77.2 & 87.3 \\
Zhou \cite{zhou2018monocap} & 68.7 & 74.8 & 67.8 & 76.4 & 76.3 & 84.0 & 70.2 & 88.0 & 113.8 & 78.0 & 98.4 & 90.1 & 62.6 & 75.1 & 73.6 & 79.9 \\
Jahangiri \cite{jahangiri2017generating} & 74.4 & 66.7 & 67.9 & 75.2 & 77.3 & 70.6 & 64.5 & 95.6 & 127.3 & 79.6 & 79.1 & 73.4 & 67.4 & 71.8 & 72.8 & 77.6 \\
Mehta \cite{mehta2017monocular} & 57.5 & 68.6 & 59.6 & 67.3 & 78.1 & 56.9 & 69.1 & 98.0 & 117.5 & 69.5 & 82.4 & 68.0 & 55.3 & 76.5 & 61.4 & 72.9 \\
Martinez \cite{martinez2017simple} & 51.8 & 56.2 & 58.1 & 59.0 & 69.5 & 55.2 & 58.1 & 74.0 & 94.6 & 62.3 & 78.4 & 59.1 & 49.5 & 65.1 & 52.4 & 62.9 \\
Fang \cite{fang2018learning} & 50.1 & 54.3 & 57.0 & 57.1 & 66.6 & 53.4 & 55.7 & 72.8 & 88.6 & 60.3 & 73.3 & 57.7 & 47.5 & 62.7 & 50.6 & 60.4 \\
Sun \cite{sun2017compositional} & 52.8 & 54.8 & 54.2 & 54.3 & 61.8 & 53.1 & 53.6 & 71.7 & 86.7 & 61.5 & 67.2 & 53.4 & 47.1  & 61.6 & 63.4 & 59.1 \\
Sun \cite{fang2018learning} & 47.5 & 47.7 & 49.5 & 50.2 & 51.4 & 43.8 & 46.4 & 58.9 & 65.7 & 49.4 & 55.8 & 47.8 & 38.9 & 49.0 & 43.8 & \textbf{49.6} \\
Moon \cite{moon2019camera} & 50.5 & 55.7 & 50.1 & 51.7 & 53.9 & 46.8 & 50.0 & 61.9 & 68.0 & 52.5 & 55.9 & 49.9 & 41.8 & 56.1 & 46.9 & 53.3 \\
\hline
\hline
ScanAva-AHuP &  135.9 & 137.2 & 104.0 & 137.1 & 139.4 & 133.6 & 140.8 & 133.7 & 163.3 & 129.5 & 137.9 & 139.5 & 123.2 & 135.1 & 130.2 &134.5 
 \\
ScanAva-AHuP PA &  75.2 & 79.1 & 68.0 & 79.1 & 91.7 & 75.8 & 82.3 & 100.9 & 128.0 & 87.4 & 83.3 & 81.8 & 76.8 & 82.2 & 78.3 &85.1 \\
\hline
Martinez (ScanAva+) \cite{martinez2017simple} &153.2 & 152.6 & 129.7 & 153.8 & 151.9 & 149.9 & 144.0 & 159.8 & 191.0 & 146.2 & 147.9 & 158.7 & 148.5 & 140.8 & 139.8 &151.5 \\
Martinez (ScanAva+) \cite{martinez2017simple} PA &97.8 & 97.9 & 93.1 & 99.5 & 105.7 & 91.3 & 91.0 & 121.4 & 139.7 & 104.6 & 96.6 & 99.5 & 105.2 & 98.2 & 98.0 &103.3 \\
\hline
\end{tabular}
}
\label{tbl:humanB}
\end{center}
 \vspace{-.1in}
\end{table*}
}

\newcommand{\tblMuPoTS}{
\begin{table*}[h]

\caption{3DPCK comparison with the state-of-the-art tested on the MuPoTS dataset. 16 out of 20 subjects are listed due to space limitation. }
 \vspace{-.1in}
\begin{center}

 \tiny
\resizebox{\textwidth}{!}{
 \begin{tabular}{ *{23}c }   
 \hline
 Methods &  S1 &  S2 &  S3 &  S4 &  S5 &  S6 &  S7 &  S8 &  S9 &  S10 &  S11 &  S12 &  S13 &  S14 & S15 & S16 & \dots &  Avg & AUC \\
 \hline
Rogez \cite{rogez2017lcr}   & 67.7 & 49.8 & 53.4 & 59.1 & 67.5 & 22.8 & 43.7 & 49.9 & 31.1 & 78.1 & 50.2 & 51.0 & 51.6 & 49.3 & 56.2 & 66.5 & \dots & 53.8 & 27.6   \\
Mehta \cite{mehta2018single} & 81.0 & 60.9 & 64.4 & 63.0 & 69.1 & 30.3 & 65.0 & 59.6 & 64.1 & 83.9 & 68.0 & 68.6 & 62.3 & 59.2 &  70.1 & 80.0 & \dots & 66.0 & 37.8 \\
Rogez \cite{rogez2019lcr}   & 87.3 & 61.9 & 67.9 & 74.6 & 78.8 & 48.9 & 58.3 & 59.7 & 78.1 & 89.5 & 69.2 & 73.8 & 66.2 & 56.0 & 74.1 & 82.1 & \dots & 70.6 &  - \\
Moon \cite{moon2019camera} & 94.4 & 77.5 & 79.0 & 81.9 & 85.3 & 72.8 & 81.9 & 75.7 & 90.2 & 90.4 & 79.2 & 79.9 & 75.1 & 72.7  & 81.1 & 89.9 & \dots &  \textbf{81.8}  &  - \\
\hline 
ScanAva-AHuP & 90.6 & 75.0 & 66.0 & 73.5 & 87.6 & 73.5 & 90.7 & 71.1 & 91.0 & 95.0 & 87.1 & 87.6 & 74.2 & 68.6 & 85.9 &  72.0 & \dots &81.6 & \textbf{39.0} \\
 \hline
\end{tabular}
}
\label{tbl:MuPoTS}
\end{center}
 \vspace{-.3in}
\end{table*}
}

\begin{document}

\title{Adapted Human Pose: \\Monocular 3D Human Pose Estimation with Zero Real 3D Pose Data}


\author{Shuangjun Liu       \and Naveen Sehgal\and
        Sarah Ostadabbas 
}


\institute{S. Liu, N. Sehgal, and S. Ostadabbas \at
              Augmented Cognition Lab, Electrical and Computer Engineering Department, Northeastern University.
              \email{shuliu@ece.neu.edu, sehgal.n@husky.neu.edu, ostadabbas@ece.neu.edu}           
}

\date{Received: date / Accepted: date}

\maketitle

\begin{abstract}
The ultimate goal for an inference model is to be robust and functional in real life applications. However, training vs. test data domain gaps often negatively affect model performance. This issue is  especially critical  for the monocular 3D human pose estimation problem, in which 3D human data is often collected in a controlled lab setting.  
In this paper, we focus on alleviating the negative effect of domain shift in both appearance and pose space for 3D human pose estimation by presenting our adapted human pose (AHuP) approach. AHuP is built upon two key components:  (1) semantically aware   adaptation (SAA) for the cross-domain feature space adaptation, and (2) skeletal pose adaptation  (SPA) for the pose space adaptation which takes only limited information from the target domain. By using zero real 3D human pose data, one of our adapted synthetic models shows comparable performance with the SOTA pose estimation models trained with large scale real 3D human datasets. The proposed SPA can be also employed independently as a light-weighted head to improve existing SOTA models in a novel context. A new 3D scan-based synthetic  human  dataset  called  ScanAva+ is also going to be publicly released with this work.\footnote{The ScanAva+ dataset is available at: \href{https://web.northeastern.edu/ostadabbas/code/}{Augmented Cognition Lab Webpage.}. The AHuP code  also can be found at GitHub \href{https://github.com/ostadabbas/AdaptedHumanPose}{AdaptedHumanPose.}}
\keywords{3D human pose estimation \and domain shift \and  semantic aware adaptation \and  synthetic human datasets.}
\end{abstract}

\section{Introduction}
\label{sec:intro}
Because of the great endeavors of the computer vision community, significant advancements have been achieved for 3D human pose estimation with ever improving performance on well recognized benchmarks \cite{h36m_pami}. 
Existing approaches come from versatile genres such as end-to-end learning \cite{li20143d}, direct 2D-to-3D lifting \cite{martinez2017simple} and even unsupervised methods \cite{chen2019unsupervised}. Although improved performance has been reported with decreased information dependence on the training sets \cite{martinez2017simple,chen2019unsupervised}, the majority of these studies are conducted via a training/testing data split from the same benchmark datasets, which share very similar contexts. 
In essence, these 3D human pose benchmarks could be quite different from the real-life applications, and the domain gap between the source data and target applications could lead to potential performance drops. 
%
And, while the domain shift issue has been extensively investigated for the classification tasks \cite{ganin2016domain,tzeng2015simultaneous}, it has rarely been addressed in the pose regression problems. 

In this paper, we investigate how domain shift influences the 3D pose estimation model  performance and, more importantly, how to counter its adverse effects. Our approach  incorporates a semantically aware adaptation (SAA) technique as well as a skeletal pose adaptation (SPA) method towards developing a robust 3D human pose estimation model. Our approach is not only capable of training a monocular 3D human pose estimation model using zero real 3D human pose data, but also can be  added to the existing state-of-the-art (SOTA) models as a light-weighted head for adaptation when tested under a novel context or dataset. 



Unfortunately, performance evaluation of the 3D pose estimation models  under  real world applications is not widely conducted, since  acquiring labeled 3D human pose data often requires professional motion capture systems \cite{h36m_pami}, and a few available datasets are collected under lab settings with very limited samples in the wild \cite{singleshotmultiperson2018}. This issue could inherently result in the similarity of the  environments, the camera settings (in both extrinsic and intrinsic parameters), and also human pose distributions between training and evaluation data splits. As a consequence, training and testing on the same benchmark disproportionately take advantage from these context similarities, which do not always hold in practical applications.  For a 2D-to-3D lifting approach \cite{martinez2017simple}, such mapping is  conducted inherently under a fixed camera setting. 
Although the model in \cite{chen2019unsupervised} employs no 3D pose data directly,  its 2D pose data is a direct projection of the 3D data from the same benchmark, which holds identical pose semantics and distributions of their target domains. 
It is also evident in their reported performance that using pose data without adaptation leads to a significant performance drop.  
Even for template-based pose estimation approaches such as \cite{bogo2016keep},  the template is  explicitly adapted to the poses from the target domain using the motion and shape (Mosh) capture  approach from sparse markers \cite{loper2014mosh}. Please note that in these approaches, to adapt the pose, the real 3D human benchmark data has to be employed in Mosh approach \cite{loper2014mosh}. 
In the recent work \cite{Li_2020_CVPR}, few 3D human training data can be extensively augmented for training and an impressive result has been achieved. Yet the focus is on augmentation, where the real 3D human data are still needed. 

Admittedly, 
it is a good practice to employ as much information as we can get for an improved performance. Yet, in a real life application, we may have access to no pose or image data in the target domain, or, at best, only limited number of measurements can be collected from the target domain a priori. 


Here, we introduce our adapted human pose (AHuP) approach, which allows us to study the monocular 3D human pose estimation problem under a highly unaware context, assuming zero  or a very  limited amount of information from the target domain. 
We conducted our study on a more challenging case, when only the  synthetic human data is employed for 3D information. With synthetic appearances, simplified skeletons, and different pose distributions, learning from synthetic humans and testing on real 3D human poses becomes a typical manifestation of the potential domain shift. 
Introducing synthetic data for learning is not a new idea,  yet competitive performance is only reported by further incorporating  real 3D human data into the current models \mbox{ \cite{varol17_surreal}}. Furthermore, there has been no specific discussion on how these differences affect the pose estimation performance and the proper way to address the challenges associated with the domain gap caused by training vs. testing data distributions. 
From our study, on one hand, with only synthetic human data for 3D part, AHuP shows comparable performance with SOTA models despite the benefit from training on real 3D human benchmark data.  
As synthetic human simulation is an efficient and easy-to-produce source pose data generator, this could potentially lead to a more economic solution for the 3D pose estimation problem. 
On the other hand, for the existing SOTA 3D human pose estimation  models, AHuP can be also added as a light-weighted adaptation head for performance improvement when a novel context or dataset is used during the test phase. This improvement is observed on several SOTA models and on all training and testing combinations in our study. 


In short, our work makes the following contributions: 

\figAHuP
\begin{itemize}
\item Presents an adapted human pose (AHuP) approach (see \figref{fig:AHuP}) that improves the performance of the state-of-the-art 3D human pose estimation models  for both synthetic-to-real and real-to-real cases when training and test data come from different contexts. By using zero real 3D human pose data, one of our adapted synthetic models shows comparable performance with the SOTA pose estimation models trained based on a large mount of  real 3D human pose data. 

\item Proposes a semantically aware adaptation (SAA) method for the cross-domain feature space adaptation, which shows noticeable pose estimation performance improvement over conventional domain adaptation techniques. 

\item Introduces a skeletal pose adaptation (SPA) approach, which takes only a limited amount of information from the target domain to adapt the pose, instead of requiring the whole 3D pose data from the target dataset for adaptation purposes, and which can be  employed independently as a light-weighted head on top of existing SOTA models for improvement under a novel context.


\item Introduces and publicly releases a new 3D scan-based synthetic human dataset, called ScanAva+, which is an extended version of our previous ScanAVA dataset (which only had 15 scans), by adding 26 new synthetic human objects with higher texture and size variations.  
\end{itemize}

\section{Related work}
In this section, we provide an overview of the state-of-the-art in the field of 3D human pose estimation and related works that leverage synthetic human models to deal with the 3D data scarcity issues. We also give a summary of the works that employ domain adaptation techniques to close the gap between their training and the test datasets, especially when  synthetic data are employed.

\subsection{The Problem of Monocular 3D Human Pose Estimation}
For monocular 3D human pose estimation, early attempts followed the 2D pose paradigm and conducted straightforward end-to-end training directly on available 3D human pose datasets such as Human 3.6M \cite{h36m_pami} and HumanEva \cite{sigal2010humaneva}. 
However, the resultant models are often not very generalizable, especially when applied on real world (``in the wild'') images.  This is mainly due to the fact that most 3D human datasets are collected under controlled lab settings that lack meaningful variations.
To improve the generalization ability in the wild,  another line of work takes a two-stage strategy, where it first trains a known 2D pose estimation model \cite{newell2016stacked}, then recovers the 3D poses on top of that \cite{zhou20153d}. Some recent works further mix 2D and 3D pose data together to solve the pose
estimation problem. 

In \cite{popa2017deep}, the authors considered 3D pose a multi-task learning process using 2D pose and depth data. \cite{zhou2017towards} further incorporated a weakly-supervised loss in order to integrate 2D and 3D data. The 3D pose estimation is also solved in a 2D-to-3D lifting manner without learning from the corresponding image or even operating in an unsupervised manner \cite{chen2019unsupervised}. 
A template-based approach has also been employed by fitting the projected silhouette with prior constraints \cite{bogo2016keep}. However, 2D-to-3D lifting or silhouette-based approaches inherently abandoned the image features, which may not always be optimal and which will be illustrated in our experiments. In recent studies, kinematic constraints have also been introduced where human motion sequence data is required \cite{Xu_2020_CVPR,pavllo:videopose3d:2019}

Another approach in 3D pose estimation is to employ synthetic human pose data, summarized  in the following section. In short, no matter what exact methodology is used, methods with competitive performance in the 3D human pose estimation field usually cannot avoid employing some real 3D human pose data in their training processes, which makes them expensive to train. 
In this work, we would like to explore this exact challenging case by using not a single real 3D human dataset in the process. 
Even though the very recent works in \cite{kocabas2019self,Remelli_2020_CVPR,Iqbal_2020_CVPR} successfully estimate 3D poses without any explicit 3D labels, they use a multi-view setting, in which 3D coordinates under that can be  estimated from the 2D pose via the multi-view geometry. However, a calibrated multi-view setting will not be always available in many applications.

\subsection{Synthetic Human Pose Data}
Employing synthetic data to solve real world problems is not a new concept in the computer vision field.  For low level vision tasks, synthetic images have been employed for stereo vision \cite{peris2012towards} and optical flow estimation \cite{butler2012naturalistic}. For higher level tasks, computer-aided design (CAD) models have also been extensively used for object detection \cite{liebelt2010multi,peng2015learning,marin2010learning}  or segmentation \cite{hong2018virtual}. 

Synthetic human figures have  been extensively used for learning purposes, such as silhouette-based action recognition tasks \cite{ragheb2008vihasi}, and crowd counting \cite{wang2019learning}.
Towards the human pose estimation,  \cite{varol17_surreal}, SURREAL provides 145 subjects and over 6.5 million frames with detailed pose and segmentation labels based on a morphable human template.
Templates from the shape completion and animation of people (SCAPE) \cite{anguelov2005scape}  have also been employed for large-scale 3D human pose dataset forming \cite{chen2016synthesizing}. 
Another branch is synthesizing human pose data directly via a human body scanning process, which could be limiting in terms of the number of scans, yet garment geometry details are better preserved, such as our recently developed Scanned Avatar (ScanAva+) dataset that contains 3D scans of 41 human subjects \cite{liu2018semi}. Nonetheless, no matter how realistic the simulated data look and despite applying conventional domain adaptation methods on them, models trained on the synthetic data alone perform noticeably worse than models trained on real human pose data \cite{chen2016synthesizing}. 


\subsection{Synthetic vs. Real Data Domain Adaptation}
Domain adaptation is a long-standing topic in the computer vision field, yet  synthetic-to-real data domain adaptation for task transfer learning  still remains a challenge. Well-known algorithms mainly address the domain gap issue at the feature level by aligning extracted features from both domains and subsequently minimizing certain distance measures between them, such as maximum mean discrepancy \cite{long2015learning}, correlation distance \cite{sun2016deep}, or adversarial discrimination loss \cite{ganin2016domain,tzeng2015simultaneous}. In recent years, domain adaptation has also been introduced for segmentation \mbox{\cite{hoffman2018cycada,luo2019taking,vu2019advent}} or object detection tasks \mbox{\cite{raj2015subspace,gopalan2011domain,chen2018domain,zheng2020cross}}, but rarely for human pose regression. 
However, such measures are usually based on the overall image features, which do not guarantee semantic consistency. Though this issue has been addressed recently \mbox{\cite{Luo_2019_CVPR}}, it usually remains a classification problem.

In the image synthesis approach described in \cite{Wang_2019_CVPR}, the authors enforce semantic consistency with an $L_1$ loss. The loss is computed per-layer between features extracted from a synthetic image and a real sampled image using a pre-trained VGG model, with an adaptive weight that decreases for deeper layers.  In \cite{10.1145/3357384.3357918}, for unsupervised domain adaptation, the authors use a two-stream convolutional neural network (CNN), one for the source domain and one for the target domain, with shared weights. When training a classifier, they assign pseudo-labels and then use an adaptive centroid alignment to offset the negative influence from false pseudo labels and enforce cross-domain class consistency. Similarly, in the segmentation task of \cite{Luo_2019_CVPR}, the authors propose a category-level adversarial network, where two classifiers identify classes whose features are distributed differently between the source and target domains and proportionally increase an adversarial loss to enforce semantic alignment across domains. 

Despite being an important concept, semantic consistency is 
commonly used for classification tasks, such as in \mbox{\cite{Luo_2019_CVPR,Morgado_2017_CVPR}}, or for image synthesis purposes \mbox{\cite{Wang_2019_CVPR}} instead of human pose.
Though adversarial learning is introduced in human pose estimation before \cite{yang20183d}, it is mainly used for pose regularization purposes, where the training portion of the real 3D human pose benchmark is required.

In this work, we  explore how 3D pose estimation performance will be negatively affected by the domain shift between source and target domains through the use of synthetic 3D human pose data as our training source. More importantly, we present the AHuP approach, which incorporates our counter action adaptation techniques in both feature space and pose space.

\section{Introducing Adapted Human Pose (AHuP)}
When a pre-trained 3D human pose estimation model is employed in a real world application, many context elements are different from the  benchmark that the model was trained on: the person will no longer wear tight clothes to facilitate the motion capture process, no tracker bead is attached, the background will be more versatile compared to a lab environment, and a different pose semantic definition may be used. 
As in most cases, human pose is estimated based on extracted image patches \cite{zhou2017towards,moon2019camera}, so we can assume the  differences  mainly come from the appearance and pose semantics. 

To investigate this problem under a more challenging scenario, we conducted our study on synthetic human since it has noticeable differences in both appearance and pose distributions compared to the common 3D pose benchmarks. Synthetic human usually holds visible unrealistic appearances with simplified skeletons, and their simulated poses, although similar, will never be the same as that of real humans. We proposed both a semantic aware adaptation approach (SAA) on the image feature space and also a skeletal pose adaptation (SPA) approach for skeleton space to align the domain gap as shown in \mbox{\figref{fig:AHuP}}. In this section, we will focus on introducing the SAA and SPA design, respectively.


\subsection{Semantic Aware Adaptation (SAA)}
\label{sec:SAA}

Domain shift is an common issue when a model learned from domain $A$ is applied in domain $B$ for a similar task. 
There are consistent efforts in the computer vision community to address such an issue. For discriminative tasks \cite{tzeng2017adversarial}, despite  variations in the specifics of the model components  \cite{hoffman2018cycada,tzeng2015simultaneous}, a commonly employed structure is to adapt two datasets $A$ and $B$  for a common task network $T$, by forming a feature extractor network $G$ (or two for an  asymmetric mapping case). 
This will map two datasets into a common feature space by minimizing a distance measure, such as the maximum mean discrepancy \cite{quinonero2008covariate}, or by confusing the discriminator with a generative adversarial network (GAN) structure \cite{tzeng2017adversarial}. 
Such distance measures are usually based on the statistics of overall feature maps \cite{tzeng2017adversarial} or local patches \cite{isola2017image}, uniformly. Although related studies are mainly conducted for classification tasks, there are also recent works that begin to address the segmentation problems with cycle consistency \cite{hoffman2018cycada}.

However, one problem often overlooked is the semantic meaning consistency between two datasets, which so far has been limited to the classification problems \cite{Luo_2019_CVPR}. 
If the same semantic entity demonstrates varying patterns in different domains, it is very possible that in an image/feature space, the nearest neighbors from two domains do not hold the same semantic meaning. A common solution to shorten the overall distance between domains is aligning the nearest neighbored patterns together to blur their domain identities. However, what if the well-aligned pattern comes out to hold different semantic meanings as \figref{fig:SAAandSPA}a displays? Such adaptation will possibly turn out to make the result even more misleading, especially for the regression tasks such as human pose estimation, where body parts are usually more similar than distinct categories in a classification task.    

\figSAAandSPA
We argue that the adaptation process should emphasize the semantic awareness in distance measures to achieve a semantically aware adaptation (SAA). 

In AHuP, an intuitive way to achieve SAA is  employing an individual discriminator for each recognizable body part. For efficiency, we use a multi-channel structure for AHuP to indicate different body parts as shown in \figref{fig:AHuP}, where $D$ stands for the discriminator, $D_i$ stands for the \textit{i}-th channel of the discriminator for corresponding joint,  $G$ stands for the feature extractor and $T$ stands for the task head as the 3D pose regression.  
As the extracted features from the backbone network $G$  suppose to be highly generalized into a low  resolution, inspired by patchGAN \cite{isola2017image}, we design the discriminator $D$ to be feature-wise by setting the  convolution kernel size to one with sigmoid. During the training process, for each channel, only the corresponding feature will be activated based on the joint location.  We avoid using any 3D human pose benchmark in this process by learning this information from the  available 2D human datasets.

We employ the adversarial learning strategy by training  $D$, $G$,  and $T$ networks  iteratively in an adversarial manner.  
For each input image $x$, we add a domain indicator $d(x)$ as 0 or 1 to indicate if it is real or synthetic, respectively.   During $D$ phase, we optimize $D$ by minimizing the  $SAA_D$ loss as: 

\begin{align}
    L_{SAA_D} &= - \sum_{i=1}^{N} \mathds{1}(c=J_{hm}(i))  
    (
    d(x)log\,D_i(G(x))[c] \\ \nonumber
    & + (1-d(x))log\,(1-D_i(G(x))[c])
    ),
\end{align}
where, $N$ is the total joint numbers, $c$ stands for the coordinate in the feature space,  $D_i(\cdot)[c]$ stands for the \textit{i}-th channel of $D$ at coordinate $c$,
$J_{hm}(i)$ is the \textit{i}-th joint location in heatmap space. Compared to the conventional adaptation approaches \cite{sun2016deep,tzeng2015simultaneous,tzeng2017adversarial}, where the feature adaptation is over the whole image region, our approach specifies the semantic meaning via multi-channel discriminator. 

We train the $G$ and $T$ networks together by minimize the regression error and confusing the $D$ at the same time.  We employ the cross entropy loss between the $D$ prediction and a uniform distribution for confusing purpose as:  
\vspace{-.3cm}
\begin{equation}
    L_{conf}= - \sum_{i=1}^{N} \mathds{1}(c=J_{hm}(i)]) (\frac{1}{2}\,\log\,D_i(G(x))[c]).
\end{equation}

The total loss during $G$ and $T$ phase is given as: 
\begin{equation}
    L_{SAA_{GT}} = ||y_{gt} - T(G(y))||_1 + \lambda L_{conf},
\end{equation} 
where $\lambda$ is the coefficient for confusion loss, $y_{gt}$ is the pose ground truth, and we use norm $L_1$ for regression supervision. 

\subsection{Skeletal Pose Adaptation (SPA)}
Pose adaptation has been extensively employed in many of the 3D human pose estimation works, especially when data outside of the benchmark domains are introduced. The main idea is to align the introduced pose or its 2D projection (\cite{chen2019unsupervised}) with 
the target domain either by direct mapping \cite{bogo2016keep} or via a discriminator to detect fake poses \cite{yang20183d}. 
The underlying assumption here is that there is extensively collected target pose data to be aligned to, which is not usually the case in the real applications. 
Instead, it is more reasonable to assume that only some countable low dimensional and interpretable parameters from the target domain could be gathered, such as the tailor measurements of the body's limbs. 

In order to achieve the skeletal pose adaptation (SPA), we use the normalized limb length vector $s(y)$ as the skeletal descriptor with the shoulder width as its normalization factor. Due to the different pose semantic definition and subjects' physiques, this descriptor could vary  among different datasets. The aligned pose is given as $ y_{SPA} = y + f(y)$, where $y$ is the output of the pose estimation network as $y=T(G(x))$ in \figref{fig:AHuP},  $f$ is a mapping function which is a multi-layer regression network with a two linear residue block in between as shown in \figref{fig:AHuP}. What we want from this mapping is a pose holding similar semantic meaning with the source domain and at the same time a skeleton similar to the target domain.  So, we employ a dual direction pivoting strategy in both pose and skeleton spaces by pushing the mapped pose to the source pose and at the same time pushing the mapped skeleton to the target skeleton, as shown in \figref{fig:SAAandSPA}b. 

To be specific, in original pose representation, by pivoting the resultant pose back to its original pose estimation $y$, we assume the mapped pose should  not be far away from the source, namely the original pose prediction, to keep the pose semantic meaning. In the skeletal representation, this is achieved by pushing the resultant pose skeleton descriptor $s(y_{SPA})$ to target $s(y_{tar})$ to make the resultant skeleton similar to target. As the skeletal descriptor $s(\cdot)$ is differentiable, our network can  effectively be updated to enforce the skeleton similarity during the model training process. 
Similar to $s(y_{SPA})$, $s(y_{tar})$ inherently comes from the target pose data $y_{tar}$. 
But as target pose data is not always available in practice, given $s(y_{tar})$ is low dimensional, we can tailor measure  $s(y_{tar})$ directly from the exact target or other subjects from the same dataset. 

The loss for SPA is given as: 
\begin{equation}
    L_{SPA} = ||f(y)||_1 + ||s(y_{SPA}) -s(y_{tar})||_2,
\end{equation}
where, $s(y_{tar})$ stands for the target skeletal measures. We employ norm $L_2$ loss for skeletal similarity pivoting and use norm $L_1$ loss for initial pose pivoting. In our design, SPA is trained after SAA, which acts as an additional component to the SAA network.  

We have to point out that though human geometric has been employed for human pose estimation \cite{zhou2017towards}, it is given as an additional constraint in an end-to-end training process where the target 3D human pose is available. 
However, SPA can be added as a light-weighted head on top of an existing pose estimation network for adaptation purposes and the training process is done without any target pose data but only with a series of low dimensional tailor measurements.  


\section{Performance Evaluation} 
To implement the AHuP approach for monocular  3D human pose estimation, we configured the architecture in \figref{fig:AHuP} by employing a ResNet \cite{he2016deep} and an integral human pose head \cite{sun2018integral} for $G$ and $T$ networks, respectively. $D$ is similar to \cite{isola2017image} in its a feature-wise manner with kernel size 1. 
\subsection{AHuP Implementation Details}
Our work is implemented via the PyTorch framework and each configuration is trained with a Nvidia v100 GPU. For the backbone feature extractor network $G$, although training from scratch converges, initialization from pre-trained weights via ImageNet can accelerate this process. Therefore, our backbone network is initialized with the pre-trained weights from ImageNet \cite{deng2009imagenet}. 
All other networks are initialized via Xavier \cite{glorot2010understanding}. 
During training and testing of the SAA network (as introduced in main text \secref{sec:SAA}), we chose batch size to be 120. 

For feature-wise $D$ shown in \figref{fig:AHuP}, we chose a 3-layer configuration with kernel size 3 stride 1. Networks $G$ and $T$ were jointly trained in an adversarial learning procedure, with $D$ serving as the counterpart. Learning rate is set at 1e-3 with a decreasing rate of 0.1 at epoch 11 and 13 with a total of 15. All input images are human-centered, cropped and resized to be 256$\times$256. The output heatmap is set as 64$\times$64$\times$64. 
During training, we initially train the 2D part at first 5 epochs to facilitate SAA process and 3D supervision is added after.

To facilitate the training and testing across different datasets, we chose the shared or similar joints to match the 17 joint configuration of the Human3.6M dataset by reordering and renaming. 
%
For data feeding, we followed a pivoted matching principle that whenever the leading feeder gives a batch, all subordinates will feed equivalent data to match. Since our study is focused on 3D human pose estimation, we always put the 3D pose dataset as the pivot feeder. For a fair comparison among varying size datasets, we fixed the iteration per epoch at 2500 by repeating the exhausted data loader, if any.  

For the SPA model, we decoupled it from the SAA net training process by learning the mapping directly from the ground truth data. 
To respect the convention of avoiding using any information from the test data, 
we used the mean skeletal descriptor $s(y_{tar})$ of the training split with the assumption that it is similar to the test split within the same dataset. 
For $f(y)$, the hidden neuron for each linear layer is set as 1024. 
The initial learning rate is set to be 1e-4 with a decreasing rate of 0.95 at each epoch with a total 70 epoch and batch size 256. 
%
During the training, the SURREAL data is downsampled with the rate of 90 to become balanced with the ScanAva+ dataset. Human3.6M is also downsampled with the rate 5 for training and 64 for testing, as commonly done in related studies \cite{moon2019camera,lassner2017unite,yasin2016dual,sun2017compositional,sun2018integral}. 
Our augmentation includes rotation, scaling, color jittering, and synthetic occlusion \cite{zhong2017random}.

\subsection{Evaluation Datasets}
For AHuP performance evaluation and comparison with the SOTA, we employed several publicly-available datasets in the human pose estimation field that have been  used extensively. For real 3D human pose data, we chose the Human3.6M \cite{h36m_pami} and MuPoTs \cite{singleshotmultiperson2018} datasets to represent lab and outdoor environments, respectively. For real 2D human pose data, we chose the MSCOCO \cite{lin2014microsoft} and MPII  \cite{andriluka20142d} datasets. As for synthetic human pose data, one branch comes from the deformable human template, in which we chose SURREAL dataset \cite{varol17_surreal}. In SURREAL, we used the released train split to extract sample images. We kept a 0.05 portion of the whole section for validation and test purposes. 

The other synthetic branch comes from the construction of virtual avatars through the direct 3D scanning of humans, in which  we chose the ScanAva+ dataset developed in our lab \cite{liu2018semi}. 
The original ScanAva has only 15 scans, which is fewer than its counterpart, SURREAL dataset. To balance the comparison in our study, we used the toolkit employed in the original ScanAva in order to collect additional human scans to augment the dataset to contain 41 full body scans and we formed the ScanAva+ dataset, in which 36 scans are used for the training. 
If not specifically indicated, we employed a basic setting with pose data from SYN + MPII + MSCOCO collectively  for all synthetic union cases, where SYN stands for either ScanAva+ or SURREAL datasets. 


\tblAbla
\subsection{Evaluation Metrics}

To provide a comprehensive view in our evaluation, we employ extensively-used metrics from real human pose benchmarks to report our performance, including mean per joint position error (MPJPE) for Human3.6M \cite{h36m_pami}, 3D percentage of correct key-points (3DPCK), and the area under curve (AUC) for MuPoTS \cite{singleshotmultiperson2018}. 
For MPJPE, we also reported the Procrustes analysis (PA MPJPE) version
\cite{gower1975generalized}, which is more reliable and fair, especially for cross-set evaluation due to varying camera parameters, joint definition, and body shape distributions. 

For 3DPCK, we follow the official configuration of \cite{singleshotmultiperson2018} with a 15cm tolerance for joint location estimation accuracy. We assume every human is correctly detected and compare all cases with pelvis rooted error. For Human3.6M, we also follow 2nd protocol during evaluation \cite{yasin2016dual,moon2019camera}, 
where subjects 9 and 11 are used for testing \cite{bogo2016keep}. Due to the joint definition differences, we use Human3.6M as a template and map the similar joints and interpolate missing ones for other datasets. 
%
Please note that in the original protocols used in the majority of the 3D pose estimation works, the training split of the Human3.6M is employed during training, which we do not use at all. This makes our task a more challenging case (and at the same time more realistic in nature) than the original protocol.

\subsection{Ablation Study}
To evaluate how the proposed AHuP framework can enhance the 3D pose estimation with only 3D synthetic data, we added each component one-by-one to form the following settings: (1) pure 3D synthetic data based learning either with SURREAL or ScanAva+, (2) learning with conventional adaptation approach by aligning the whole feature space directly similar to  \cite{chen2016synthesizing,long2015learning,sun2016deep}, named with suffix 'C', (3) semantic aware adaptation with suffix 'SAA', (4) further adding 2D pose task from additional 2D human pose dataset (MSCOCO and MPII datasets) with suffix 'Jo2D', (5) further adding  skeletal pose adaptation with suffix 'SPA'.

From the results shown in \tabref{tbl:abla}, we can see that conventional adaptation by aligning the whole features \cite{tzeng2017adversarial,tzeng2015simultaneous} does improve the performance on  both synthetic datasets across all real benchmarks.  
By employing the SAA strategy, the improvement is significant on ScanAva+, and  slightly but still noticeable on SURREAL.  
Additional 2D pose tasks from the real 2D human dataset shows further improvement, which agrees with the existing studies \cite{zhou2017towards}. SPA shows noticeable improvement for both ScanAva+ and SURREAL on Human3.6M but not much difference on MuPoTS. Although MuCo is  the training split for MuPoTS  \cite{singleshotmultiperson2018}, the two datasets are in fact captured in different environments separately,
which could result in differences in the skeletal descriptors. 
We still notice that in the SURREAL case, although the improvement is not obvious on PA MPJPE, the 3DPCK metric is improved significantly. 
It shows that by adding SPA , many of the pose errors fall back into the tolerance range.

\subsection{Qualitative Study in Pose and Feature Spaces}
\tabref{tbl:abla} shows that applying the AHuP approach leads to pose estimation performance improvement on both real 3D pose benchmarks; however, there are much stronger improvements when ScanAva+ dataset is used for training compared to the SURREAL.  To figure out the underlying reason, we further investigated the characteristics of the datasets themselves. We randomly extracted 5000 3D human pose samples from all four 3D datasets, including real (Human3.6M and MuPoTS) and synthetic (ScanAva+ and SURREAL) pose datasets, and visualized them via a t-distributed stochastic neighbor embedding (t-SNE) approach \cite{maaten2008visualizing} in \figref{fig:tSNEpose}. 
This plot is purely based on the raw pelvis rooted 3D pose data 
without filtering after matching the joint order across datasets, 
in order to reflect the essential pose difference among these sets. From the plot, surprisingly we found out a higher agreement between all real datasets and ScanAva, yet a clear boundary around SURREAL. 
It seems SURREAL does not hold a well-overlapping pose manifold with the others. 
The causes could be multi-fold,  including  the camera setting and joint definition in their rendering process, which all  possibly affect the final pose distribution.  When SURREAL is the only 3D source, the model can hardly learn any more than its pose coverage. This presumes to be the cause of limited improvements in the case when only SURREAL data is used. 


\figTSNEpose

\figTSNEapr

It is also interesting to investigate how these approaches influence the feature space. To better illustrate this, we visualized the output features of $G$ from all four datasets under different model configurations as shown in \figref{fig:tSNEapr}.
Due to computational intensity and also to prevent a cluttered  visualization, we evenly sampled 100 images from each dataset, with the $G$ network's output features downsampled to $1024\times4 \times4$. 
\figref{fig:tSNEapr} shows how differently each dataset is in the ``eye'' of these models. In \figref{fig:D0}, all datasets show clear clustering effect. 
This is reasonable as the model based on only synthetic data can hardly generalize well to capture the shared features for both of real and synthetic data. 
With the adaptor introduction in \figref{fig:C}, the clustering effect has been eliminated much, but we notice that the synthetic features are more located on the right hand side for both SURREAL and ScanAva+. With SAA as shown in \figref{fig:SAA}, there is no obvious improvement over the C version, but the synthetic features are more evenly distributed in the space. With additional 2D task from real 2D human pose in \figref{fig:SAA-y}, the distribution becomes flattened. One interesting observation is that SURREAL shows a more obvious clustering effect than the adaptation-only version. 
Instead of losing the generalization ability, we believe this clustering effect on the contrary indicates an improved ability of the network to recognize different poses.  As SURREAL does not hold a similar pose distribution to the others, it supposes to show a different pattern in a well recognized pose space.

\subsection{Adaptation of SOTA Models using AHuP}
Although our proposed AHuP approach shows improvement over the models trained on the synthetic human data, 
one immediate question is why bother to use AHuP given that well-performed 3D human pose models already exist? 
Here, we examine how AHuP is also capable of improving the performance of the existing SOTA 3D pose estimation models when tested under a different context or dataset from their training set.

In order to conduct a fair comparison, we designed a cross-set evaluation experiment that all candidate models are trained and tested on different datasets to mimic the effect when they are employed in applications under a novel context.
We chose two SOTA pose estimation models introduced by Sun et al. in \cite{sun2018integral} and by Zhou et al. in \cite{zhou2017towards} in this experimental analysis. To cover more possible scenarios, for top performer \cite{sun2018integral}, we trained two networks with Human3.6M + MPII + MSCOCO  and  MuCo + MPII + MSCOCO, respectively.
For \cite{zhou2017towards}, we employed their official release of the pre-trained model on Human3.6M + MPII. 
We tested both model on a 3D human pose dataset that was novel for them. As they already learned from the real domains with limited shift in the appearance, we only evaluated the effect of our SPA on these models. 
Their model performance with or without SPA is reported in \tabref{tbl:SPA}. The results illustrate that adding AHuP adaptation in the form of SPA to these models leads to consistent performance improvement over the versions without SPA. These improvements are seen  in both models \cite{sun2018integral,zhou2017towards},in all of  the  training set combinations, and for different   types of  test sets.


\subsection{Practical Values of AHuP}
SPA can be employed as a switchable head on top of an existing network such as ResNet to act as a light-weighted adaptation strategy.
The adaptation can be achieved by simply switching the SPA head without retraining the network itself. The benefits of this SPA adaptation strategy include: 

\begin{itemize}

     \item Time efficient adaptation: To evaluate a model on different benchmarks, it is common to train a specific model for each of the new benchmarks, as suggested by \cite{moon2019camera}. However, retraining the model in \cite{sun2018integral}  on a new dataset takes two days \cite{moon2019camera}. In contrast, training a SPA head takes less than 20 minutes. 

     \item Memory efficient storage: Suppose we train a new model for each new dataset/context; the storage cost will be proportional to the number of the datasets scaled by the size of the network. However, in SPA adaptation, we only need one copy of the model (e.g., PoseNet) with different SPA adaptation heads. An example of the storage comparison for models in \cite{sun2018integral,zhou2017towards} with and without SPA strategy is shown in \figref{fig:SPAparams}, where the more potential datasets there are to work on, the more memory saving our SPA adaptation it will lead to.

    \item  Efficient computational cost: Adding SPA in inference processes will inevitably increase the calculation cost, but the cost increase is actually negligible compared to the computation of the original network.  A comparison of computational cost of the two models in \cite{sun2018integral,zhou2017towards} is shown in \tabref{tbl:SPAflops}. 
\end{itemize}

\tblSPA
\figSPAparams
\tblSPAflops

\tblHumanB
\tblMuPoTS 
\figAAICvis

\subsection{Comparing 3D Pose Estimation Performance of AHuP with the SOTA}
Fundamentally, our  AHuP design focuses on performance improvement under a cross-set evaluation scenario.  Nonetheless, to show AHuP performance in a larger picture, we directly compare AHuP model performance trained on (ScanAva+ + MSCOCO + MPII) with the SOTA models  on the reported metrics as shown in \tabref{tbl:humanB}.
Please note that this  comparison is between AHuP without using any real 3D human pose data, and other models  which  directly benefit from using the training and test split from the same benchmarks \cite{h36m_pami,singleshotmultiperson2018}, with no domain shift to overcome at all \cite{sun2018integral,yasin2016dual,moon2019camera}. 
Following the convention in \cite{moon2019camera}, we reported our performance under Human3.6M Protocol\#2 as shown in \tabref{tbl:humanB}. As 3D pose estimation is a scale-uncertain process, different datasets have different camera poses and parameters, 
which will directly affect the  regression results. So we also included the  rigid Procrustes analysis (PA) alignment \cite{gower1975generalized} result for a  more just comparison. 
Though it cannot match the best performing models, we demonstrate that our model can already rival some approaches that learn directly from real 3D human pose data  \cite{yasin2016dual,chen20173d}. 
Furthermore, the difference in pose definition could  introduce additional estimation errors.
For example, in ScanAva+ the ``head'' joint lies at the top point of the head, yet in Human3.6M, this joint is biased towards the head center. The ``ankle joint'' in Human 3.6M lies at the back of the heel, yet ScanAva+ and SURREAL place them closer to the center of the ankle above the foot. 

One question is whether or not  we really benefit from the learned features that are extracted from the synthetic data with the SAA, or does it only constitute a 2D-to-3D lifting. To investigate this question, we specifically trained a  2D-to-3D lifting SOTA model \cite{martinez2017simple} under the same setting of AHuP with synthetic ScanAva+ data. Its performance is reported as Martinez (ScanAva+) in the \tabref{tbl:humanB}, which shows that AHuP still performs noticeably better than 2D-to-3D lifting version when employed under the same setting.  


Another set of real 3D pose data evaluations is conducted on MuPoTS \cite{singleshotmultiperson2018} with results shown in \tabref{tbl:MuPoTS}. It is quite surprising that despite the fact that we do not use a single frame of real 3D human data, AHuP shows competitive performance among SOTA models that are trained on the  real 3D human pose data.  
This mainly stems from the fact that MuPoTS is collected in the wild with multiple people, so the images are in a more natural setting. In comparison, Human3.6M is collected in a studio environment with a limited number of subjects.
For a fixed lab setting such as Human3.6M, many context factors, such as camera pose and background, can be  inherently well-studied from its corresponding training set, but it is not the case for MuPoTS, since its data seems to have higher variations in these factors. 


\subsection{Qualitative Comparison}
We also visualized the recovered 3D pose results when AHuP trained on ScanAva+ is used (see \figref{fig:AAICvis}). Despite the lower performance compared to the top rank 3D pose estimation models, the  recovered skeletons via our method agree well with human perception. 
In fact, when recovered joints are within a tolerable error threshold, from a human perspective, our prediction is ``semantically'' correct. 
We also  tested AHuP on synthetic data in the last two columns of \figref{fig:AAICvis}, which performs equivalently well. 

\subsection{Evaluation in 3D Multi-Person Pose Estimation}
Our approach also shows compatibility with other works. Combining with the proposed approach in \cite{moon2019camera} by employing Mask-RCNN \cite{he2017mask} as detect-net (for human detection), our model can be well integrated as a pose-net (for 3D human pose estimation) for multi-person pose estimation. As multi-person performance is also affected by root localization, we only report qualitative results, as shown in \figref{fig:AAICmulti}  without taking credits from \cite{moon2019camera}. 

\figAAICmulti

\section{Conclusion}
The ultimate goal of training an inference model is to make the model ready to perform in some real world applications. Training and testing under different contexts  potentially introduces the domain gap and influences the model performance negatively. This issue is especially magnified in the  3D human pose problem, where the majority of the 3D human pose benchmarks are collected under controlled lab settings. 
To mitigate this effect, we presented our adapted human pose (AHuP) approach that incorporates a semantic awareness adaptation (SAA) technique as well as a skeletal pose adaptation (SPA) algorithm and illustrated how AHuP improves 3D pose estimation model performance both quantitatively and qualitatively. 
For a better illustration of an application with a significant context shift, we chose the synthetic human data to train our inference model without using any real 3D human pose data. We then  tested AHuP on the well-known  3D human pose benchmarks, in which it  showed comparable performance with many of the state-of-the-art (SOTA) models, which have full access to the real 3D pose data.
For existing SOTA models, our approach can  also be added as a light-weighted adaptation head which showed consistent improvement for all the candidate models, over all training and testing combinations in our study.   
Admittedly, without having access to the target real 3D human data, AHuP has a challenge in beating the best performers. However,  in a real-life problem, the  solution we care most about is often not what we can achieve under an ideal  condition (such as a controlled lab setting) but how well we can get a solution when given limited access to target data under the practical constraints. 



\bibliographystyle{spmpsci}
\bibliography{paper}

\end{document}